\documentclass[runningheads]{llncs}
\usepackage{graphicx}

\usepackage{tikz}
\usepackage{comment}
\usepackage{amsmath,amssymb} 
\usepackage{color}

\usepackage{multirow}

\usepackage{pifont}
\newcommand{\cmark}{\ding{51}}%
\newcommand{\xmark}{\ding{55}}%

\usepackage[accsupp]{axessibility}  

\begin{document}
\pagestyle{headings}
\mainmatter

\title{Beyond Visual Field of View: Perceiving 3D Environment with Echoes and Vision}

\titlerunning{Beyond Visual Field of View}
%
\author{Lingyu Zhu\inst{1} \and
Esa Rahtu\inst{1}
\and
Hang Zhao\inst{2}
}
\authorrunning{L. Zhu et al.}
%
\institute{Tampere University\\
\email{lingyu.zhu@tuni.fi}, \email{esa.rahtu@tuni.fi}\\
\and
Tsinghua University\\
\email{zhaohang0124@gmail.com}}
\maketitle

\begin{abstract}

This paper focuses on perceiving and navigating 3D environments using echoes and RGB image. In particular, we perform depth estimation by fusing RGB image with echoes, received from multiple orientations. Unlike previous works, we go beyond the field of view of the RGB and estimate dense depth maps for substantially larger parts of the environment. We show that the echoes provide holistic and in-expensive information about the 3D structures complementing the RGB image. Moreover, we study how echoes and the wide field-of-view depth maps can be utilised in robot navigation. We compare the proposed methods against recent baselines using two sets of challenging realistic 3D environments: $\textit{Replica}$ and $\textit{Matterport3D}$. The implementation and pre-trained models will be made publicly available.

\end{abstract}

\section{Introduction}

The structure of a 3D environment is commonly inferred using RGB images or active depth sensors. While they provide detailed information, the observations are usually limited to a small field of view. This limitation can be compensated by installing multiple cameras or physically moving the device. However, such procedure requires processing multiple images, which might be unnecessarily heavy for completing the required tasks (e.g.~ navigation).

Several animal species, such as bats, dolphins, and some nocturnal birds, perceive spatial layout and locate objects through echolocation~\cite{rosenblum2000echolocating,gao2020visualechoes,parida2021beyond}. By using two ears to receive spatial sound, one can determine the objects' location by the Interaural Time Difference (ITD) and Interaural Level Difference (ILD). The received sound is a function of the room geometry, surface materials~\cite{antonacci2012inference}, and the receiver's location. Despite echoes and bat-like echolocation approaches~\cite{eliakim2018fully,antonacci2012inference,Dokmani2013AcousticER,vanderelst2015sensorimotor,frank2020comparing,christensen2020batvision} have been proposed for perceiving spatial layout of different environments, the audio-based spatial reasoning still remains a challenging problem.

\begin{figure}[!tbp]
    \centering
    \includegraphics[width=1.0\textwidth,keepaspectratio]{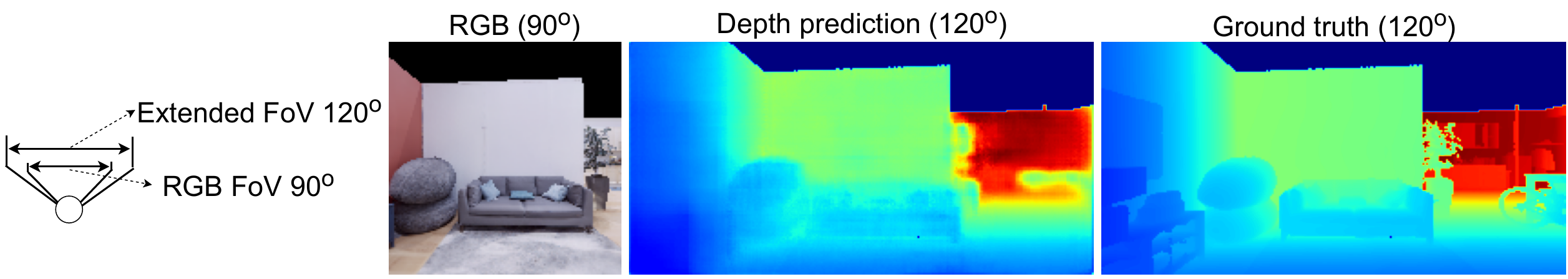}
    \caption{Leveraging echoes to extend depth prediction over RGB FoV.}
    \label{fig:fig1}
\end{figure}

Recent works~\cite{gao2020visualechoes,parida2021beyond} have shown that jointly utilizing the audio-visual observations can substantially enhance the spatial reasoning of physical space. However, these methods mostly focus on enhancing the prediction in the same area as the RGB covers. The improvement from leveraging audio signal to RGB is rather limited when staying inside the RGB field of view (FoV). One advantage of echoes is that echoes naturally have a wider ``field of view '' than the RGB observation. Instead of focusing on inside the RGB FoV, one can benefit most from outside the RGB FoV when leveraging echoes. These motivate us to study taking advantage of echoes information to go beyond the RGB FoV and to extend the prediction for wider FoV (see Fig.~\ref{fig:fig1}). For instance, one has access to an RGB image of FoV 90$^{\circ}$, a system with echoes can extend the prediction over RGB FoV.

Depth contains information relating to the distance of the surfaces of scene objects from a viewpoint. It has brought strong spatial cue of physical space to embodied agent for facilitating obstacles avoidance and navigation in 3D environment~\cite{chen2020soundspaces,chen2020learning,yu2021sound}. In this paper, we first study how to estimate a wide FoV depth from leveraging echoes with RGB and then apply them for embodied 3D navigation. 

In embodied AI, the navigation task has tightly integrated seeing and moving for numerous applications of rescuing, searching, and service robotics. Very recently, researchers incorporate acoustics~\cite{chen2020soundspaces,gan2020look,chen2020learning,chen2021semantic} to complement the visual information. In~\cite{chen2020soundspaces,gan2020look,yu2021sound}, the embodied agent is required to navigate to a sound-emitting target using the received egocentric audio and vision. In contrast, we introduce PointGoal echo navigation for complex and realistic 3D environments. We let the agent keep emitting and receiving audio signal while moving in the 3D environments. This does not require the environment to offer a sound-emitting source, which is a more realistic and practical scenario. Our approach sheds lights on directly utilizing binaural echoes to have a holistic understanding of the environment for better navigation. We show that using echo to perceive physical space for navigation outperforms using RGB. Incorporating echoes with visual observation further improves the navigation performance.

In summary, our key contributions include: i) an end-to-end neural network architecture that learns to take advantage of echoes received from multiple orientations and RGB image for better depth estimation; ii) leveraging echoes to extend depth prediction over RGB FoV; iii) using the depth, predicted from echoes and RGB, as the input for 3D navigation outperforms the result from using raw RGB as the input; iv) introducing a novel PointGoal echo navigation that directly utilizing binaural echoes to holistically perceive the physical space of realistic 3D environments. It outperforms the method of using RGB. v) Fusing the echoes to visual observations further improves the navigation performance. Without adding more cameras and additional processing, utilizing echoes helps to overcome the limitations of narrow visual FoV and to obtain better understanding of the 3D environment.

\section{Related Work}

\paragraph{\bf Audio-Visual Learning:}
Recent research bridges the audio and vision for various cross-model learning tasks. Some have achieved remarkable performance in audio-visual action recognition~\cite{kazakos2019epic,gao2020listen,lee2020cross}, audio-visual correspondence~\cite{aytar2016soundnet,arandjelovic2017look,arandjelovic2018objects}, audio-visual synchronization~\cite{owens2018audio,korbar2018cooperative,zhu2022visually}, visual sound separation~\cite{ephrat2018looking,zhao2018sound,zhao2019sound,zhu2020visually,gao2021visualvoice,zhu2021leveraging,tian2021cyclic,zhu2022v}, visual to auditory~\cite{zhou2018visual,gan2020foley,zhou2020sep,rachavarapu2021localize,xu2021visually}, audio spatialisation~\cite{christensen2020batvision,yang2020telling,morgado2020learning,zhou2020sep,gao2020visualechoes,rachavarapu2021localize,xu2021visually,parida2021beyond}, and audio-visual navigation~\cite{chen2019audio,chen2020soundspaces,chen2020learning,gan2020look,dean2020see,chen2021semantic,majumder2021move2hear}. In this work, we leverage audio-visual learning for better perceiving the geometrical structure of environment.

\paragraph{\bf Spatial Reasoning with Echoes:}

Previous work of~\cite{antonacci2012inference,Dokmani2013AcousticER,frank2020comparing} has studied using echo for reasoning object shape and room geometry of the environment. A recently proposed low-cost BatVision~\cite{christensen2020batvision} estimated the 3D spatial layout of space ahead by just listening with two ears. Bat-like echolocation approaches have been applied to facilitate autonomous robots to avoid obstacles~\cite{vanderelst2015sensorimotor}, map and navigate~\cite{eliakim2018fully} in dynamic environments.

Ye~\textit{et al.}~\cite{ye20153d} and Kim~\textit{et al.}~\cite{kim20173d} have shown that the echoes can be a strong complementary source for vision, especially when the visual information is unreliable. Recently proposed works by Gao~\textit{et al.}~\cite{gao2020visualechoes} and Parida~\textit{et al.}~\cite{parida2021beyond} are most related to ours. \cite{gao2020visualechoes} learned visual representations via echolocation for multiple downstream vision tasks. In~\cite{parida2021beyond}, the authors utilized RGB images, binaural echoes and pretrained material properties of objects to estimate depth. In contrast, our work looks into echoes received from different orientations and offers a new perspective on using echoes to overcome the limited FoV of visual observations. In addition, we propose PointGoal echo navigation to directly use echoes for perceiving a holistic understanding of environment in navigation problem.

\paragraph{\bf Monocular Depth Estimation:}

Monocular depth estimation has been recently shifted to improving neural network architectures and optimizing methods~\cite{liu2015deep,wang2015towards,xu2017multi,fu2018deep,bhat2021adabins,huynh2021monocular,miangoleh2021boosting,Yang_2021_ICCV}, integrating hierarchical features~\cite{xu2017multi,lee2019big,miangoleh2021boosting}, leveraging camera motion between pairs of frames~\cite{zhou2017unsupervised,jiang2018self,Godard_2019_ICCV,ranjan2019competitive}, taking advantage of planner guidance~\cite{lee2019big,huynh2020guiding} and 3D geometric constraints~\cite{mahjourian2018unsupervised,chen2019self,luo2020consistent,huynh2021boosting}. More recently, audio~\cite{vasudevan2020semantic,gao2020visualechoes,parida2021beyond} has been introduced to help for estimating depth. These approaches mainly focus on estimating the depth inside the RGB FoV. Unlikely, our work addresses the FoV limitation problem of RGB and concentrates on leveraging echoes to estimate the depth outside the RGB FoV.

\paragraph{\bf Learning to Navigate in 3D Environments:}

Realistic 3D environments and simulation platforms~\cite{ammirato2017dataset,chang2017matterport3d,xia2018gibson,straub2019replica} greatly facilitate embodied AI agents for the visual navigation research. Recently, AI agents are often equipped with egocentric visual observations~\cite{zhu2017target,zhu2017visual,Mirowski2017LearningTN,jain2020cordial,chaplot2020learning,chaplot2020neural} for performing question answering~\cite{gordon2018iqa,das2018neural,Thomason2019VisionandDialogN,das2020probing}, instructions following~\cite{Anderson_2018_CVPR,Chen_2019_CVPR,anderson2021sim}, and active visual tracking~\cite{luo2018end,zhong2021advat,zhong2021distraction}. Previous studies have shown audio as a strong cue for obstacle avoidance and navigation~\cite{thinus1997representation,gunther2004using,fortin2008wayfinding,picinali2014exploration,massiceti2018stereosonic,evers2018acoustic,woubie2019autonomous}. However, these methods are either non-photorealistic, not supporting AI agents, or rendering audio not geometrically and acoustically correct. 

The emergence of SoundSpaces~\cite{chen2020soundspaces} enables realistic audio-visual rendering for AI agent training in complex 3D environments. Chen~\textit{et al.}~\cite{chen2020soundspaces} and Gan~\textit{et al.}~\cite{gan2020look} introduced AudioGoal that requires the agent navigate to a sounding goal using audio and vision. Moreover, recent studies have developed a line of audio-visual navigation models~\cite{chen2019audio,chen2020learning,majumder2021move2hear,chen2021semantic,yu2021sound}. Unlike existing methods heavily relying on visual observations and sounding targets, we propose PointGoal echo navigation which sheds lights on utilizing the holistic geometry contained in binaural echoes (emit and receive sound by agent) for efficient navigation. We further fuse echoes into visual observations for better navigation performance.

\begin{figure}[!tbp]
    \centering
    \includegraphics[width=0.8\textwidth,keepaspectratio]{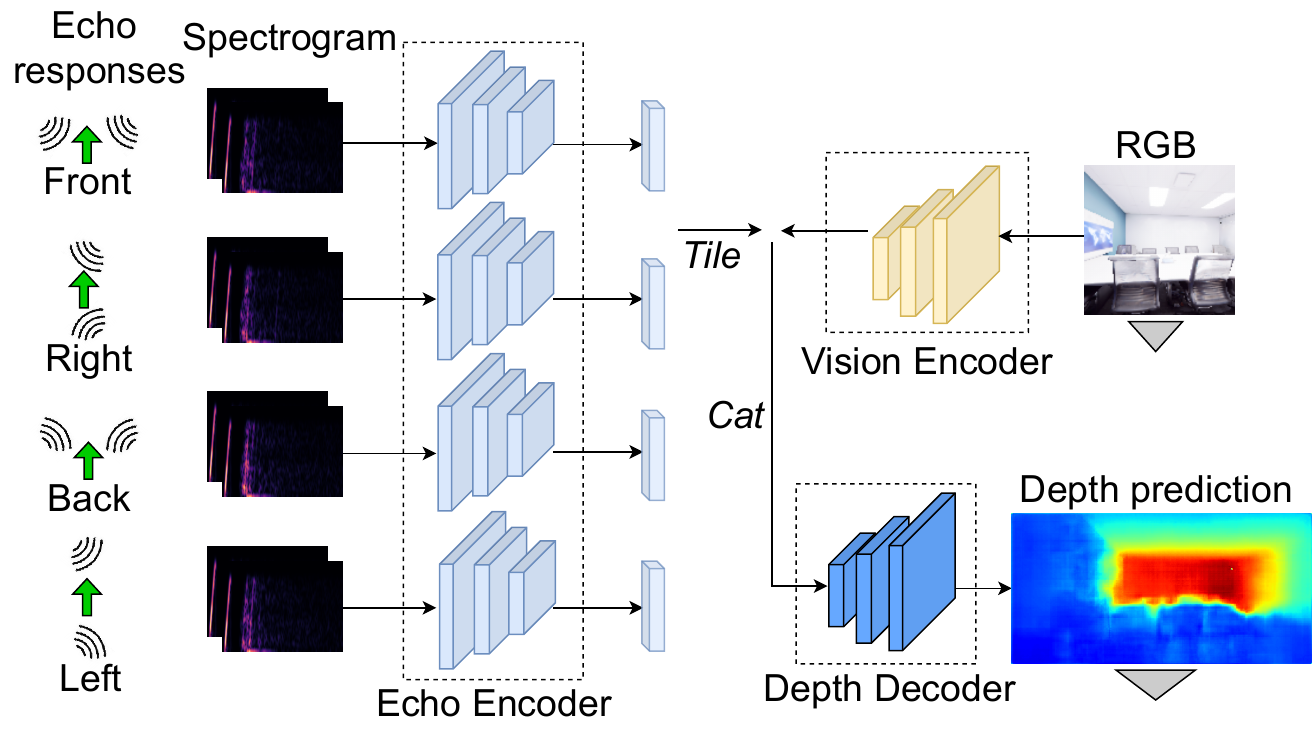}
    \caption{The framework of depth estimation from echoes and RGB image. It consists of four echo encoders, a vision encoder, and a depth decoder. The four echo responses represent the echoes received from four different orientations (e.g., front, right, back, left side) relative to the green ``arrow'' (target depth orientation). The shaded gray triangle represents the field of view for the RGB input or predicted depth. }
    \label{fig:arch_depth}
\end{figure}

\section{Predicting Wide Field of View Depth Maps from Echoes and RGB}
\label{sec:echo_depth}

\subsection{Overview}
In this section, we study how to estimate a large field of view depth map using a narrow field of view RGB and echoes, received from multiple orientations. Given only echoes as the input, we propose a model in Fig.~\ref{fig:arch_depth} (without the vision encoder) to utilize signals received from different orientations to predict the depth. Each echo encoder maps a pair of binaural echo spectrograms into a vector, which reserves the spatial cues of the 3D environment. The computed spatial vectors from different echo orientations are concatenated first before passing to the depth decoder for depth prediction. The echo encoders share parameters. When an RGB with echoes is available, we implement the whole model in Fig.~\ref{fig:arch_depth} to fuse echoes into RGB for predicting a wide FoV depth.

\subsection{Depth Estimation From Echoes}
\label{sec:depth_echo}




In order to take advantage of echoes from multiple orientations, we propose a framework in Fig.~\ref{fig:arch_depth} (omitting the vision encoder). It contains four echo encoders and a depth predictor. The echo encoder is a convolutional neural network. It is composed of three continuous blocks of $\{\mathit{Conv}, \mathit{BatchNorm}, \mathit{ReLU}\}$. A convolution layer with kernel $1\times1$ is appended at the end to convert the echo features into a vector of size $512 \times 1$. The input of each echo encoder is represented as two-channel Frequency-Time (F-T) spectrograms that are obtained by applying Short-time Fourier Transform (STFT) to the received binaural echo responses. The depth decoder consists of 6 blocks of $\{\mathit{ConvTranspose}, \mathit{BatchNorm}, \mathit{ReLU}\}$ and a following $\{\mathit{ConvTranspose}, \mathit{Sigmoid}\}$ for projecting features into one channel depth prediction within the range of [0, 1]. We provide further details in supplementary materials.

\subsection{Depth Estimation from Echoes and RGB}
\label{sec:depth_av}

RGB image is a strong cue for inferring the depth. However, the RGB is often available only for very limited FoV and thus provides only narrow picture of the scene when considering the human-like setup. Adding more cameras will introduce lots of additional processing. However, the ambisonic audio received from the omnidirectional signal is naturally a sensory signal that is equivalent to ``360$^{\circ}$ image'', which provides rich holistic geometry information of the 3D environment. Echo could be a very strong cue when we go outside the RGB FoV. These motivate us to leverage echoes to overcome the limitations of visual observation and obtain better perception of the environment.

\subsubsection{Estimating depth maps beyond the visual field of view: }
The architecture in Fig.~\ref{fig:arch_depth} fuses the echoes into RGB for predicting a wide FoV depth. The vision encoder consists of five convolutional layers. Each layer is followed by a $\mathit{BatchNorm}$ and $\mathit{ReLU}$. We tile the encoded echo feature vector to match the spatial dimension of visual features, and then concatenate the echo and visual feature maps along the channel dimension to pass to the depth decoder. 

Given an RGB image with FoV $\theta$, e.g., $120^{\circ}$, we augment an input RGB image of a smaller FoV $\theta^{'}$ by masking an ``unseen" region from two sides of this full RGB image as zeros. The new $width^{'}$ corresponding to the FoV $\theta^{'}$ is computed through,
\begin{equation}
    \centering
    \label{eq:2}
        width^{'} = width * tan(\frac{\theta^{'} * \pi}{360^{\circ}})/tan(\frac{\theta * \pi}{360^{\circ}}),
\end{equation}
where the $width$ represents the original RGB width corresponding to a full FoV $\theta$. The $width^{'}$ is the computed RGB width corresponds to a specific smaller FoV $\theta^{'} \in (0, \theta]$. Note that we only consider the FoV changes in horizon, and 360$^{\circ}$ is the full degree of a horizontal plane. We provide more details in the supplementary materials.

\subsubsection{Extending depth prediction to complete unseen areas: } 

One might be interested in the information form the unseen areas, for instance, predicting depth of the ``left'', ``right'', or ``back'' side when facing forward. Due to the visual similarity and extension of the environment surfaces, RGB image may potentially benefit for inferring the depth of unseen areas. Here we study a more extreme case, that is to predict the depth of a totally side way or opposite way when giving echoes and RGB image. We take echoes and RGB image (FoV of 90$^{\circ}$) as the input to the architecture in Fig.~\ref{fig:arch_depth}. Note that there is no overlap among the RGB observation received from ``front'', ``left'', ``right'', or ``back''.

\begin{figure}[!tbp]
    \centering
    \includegraphics[width=0.8\textwidth,keepaspectratio]{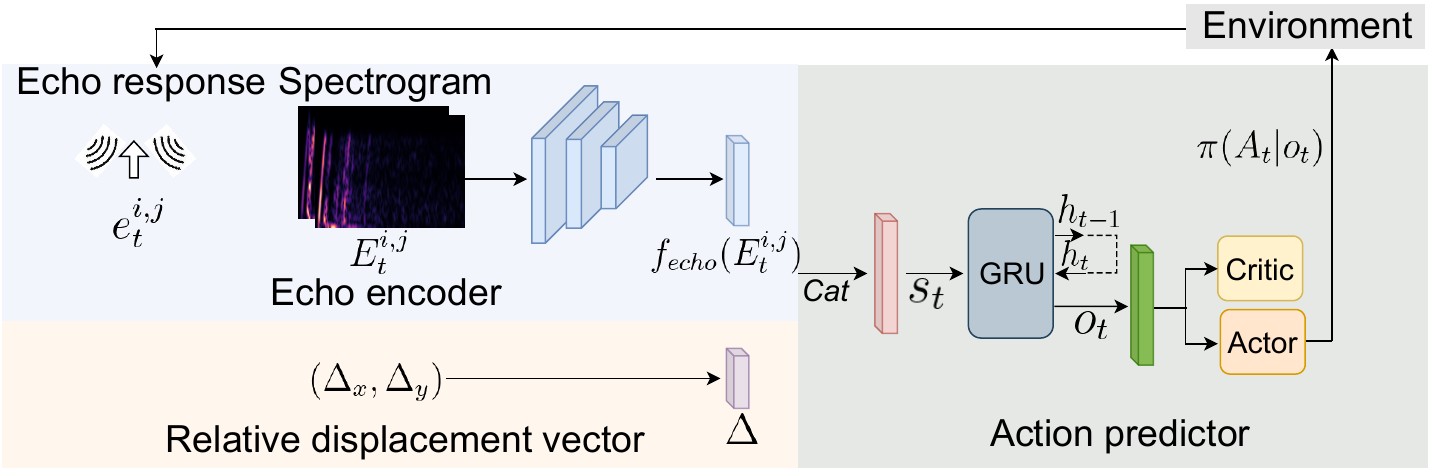}
    \caption{The architecture of the PointGoal echo navigation.}
    \label{fig:arch_nav}
\end{figure}

\section{Navigating Using Echoes and RGB}
\label{sec:echo_nav}

\subsection{Overview}

We introduce PointGoal echo navigation (Fig.~\ref{fig:arch_nav}) to directly use echoes to perceive the spatial cues of physical space for 3D navigation. The echo navigation network is composed of an echo encoder and action predictor. The echo encoder maps the binaural echoes into a vector. The action predictor processes the echo feature vector and GPS signal to predict agent actions. Moreover, we take advantage of audio-visual learning by fusing echoes to visual observations for better embodied 3D navigation.

\subsection{PointGoal Echo Navigation Task Setup}

In PointGoal navigation, a randomly initialized agent utilizes sensory inputs to navigate, avoid obstacles, and reach a given target point goal. PointGoal visual navigation is often studied for the navigation task. It considers visual sensory inputs of raw RGB or depth images to perceive the spatial information within the agent's egocentric view. Visual observation is often available only for very limited FoV and thus provides only narrow picture of the scene. However, binaural echoes contain holistic geometrical information of the 3D environment, which motivates us to directly use binaural echoes for the navigation tasks.

We introduce PointGoal echo navigation to directly utilize binaural echoes for embodied 3D navigation. Given a point goal defined by a displacement vector $(\Delta_{x}, \Delta_{y})$ relative to the randomly initialized starting position of the agent, the task of PointGoal echo navigation is to let the agent navigate to the point goal by keeping receiving binaural echoes while moving. Note that there is no map of the scene is available to the agent. The agent needs to avoid obstacles, navigate, and reach the target goal by perceiving spatial cues using sensory input. The sensory inputs are GPS and binaural echoes. An idealized GPS sensor~\cite{savva2019habitat,kojima2019learn,gordon2019splitnet,Chaplot2020LearningTE} offers the relative location of the target goal. To emit the omnidirectional sweep signal and receive the binaural echoes, we emulate one speaker and two microphones on the agent. The navigation actions consist of four actuations: \textit{MoveForward}, \textit{TurnLeft}, \textit{TurnRight}, and \textit{Stop}. More details regarding the agent embodiment and action space are provided in the supplementary materials.

\subsection{Echo Navigation Network}

We introduce echoes to input observations, and study a policy for mapping the sensory input to agent actions by adopting deep reinforcement learning with Proximal Policy Optimization (PPO)~\cite{schulman2017proximal} in Fig.~\ref{fig:arch_nav}.

\paragraph{Echo Encoder:} 

Similar as the echo encoder introduced in Section~\ref{sec:echo_depth}, the echo encoder we adopt for the PointGoal echo navigation task is also a convolutional neural network which stacks three convolutional layers. Differently, for better preserving the geometrical information, we remove the $\mathit{BatchNorm}$~\cite{ioffe2015batch} layer before applying the nonlinear $\mathit{ReLU}$ operation. We apply STFT on top of the received binaural echo responses $e^{i,j}_{t}$ to compute the binaural echo spectrograms $E^{i,j}_{t}$ as the input of the echo encoder $f_{echo}$. The $e^{i,j}_{t}$ and $E^{i,j}_{t}$ indicate, at time step $t$, the binaural echo response and its spectrograms of the agent at location $i$ with orientation $j$. The following flatten and fully connected layer map the echo features into a feature vector of size $512 \times 1$, which preserves the room geometry and agent's position.

\paragraph{Action Predictor:} 

The echo features $f_{echo}(E^{i,j}_{t})$ carry important spatial information and make the agent aware of its position and the room geometry, which is beneficial for the agent to execute an action towards the point goal. The Gated Recurrent Unit (GRU)~\cite{cho2014learning,chung2015recurrent} module takes as input the concatenation of the echo feature vector $f_{echo}(E^{i,j}_{t})$ and the given GPS displacement vector $\Delta$ to recursively process each symbol while maintaining its internal hidden state $h$. The following is a reinforcement learning policy equipped with an actor-critic architecture. It produces a probability distribution $\pi(A_{t}|o_{t})$ over possible actions by operating on the predicted agent state $o_{t}$ from the GRU module,
\begin{align}
    \centering
    \label{eq:3}
        s_{t} &= \textit{Cat}(f_{echo}(E^{i,j}_{t}), \Delta),\\
        &o_{t}, h_{t} = f_{gru}(s_{t}, h_{t-1})
\end{align}
where the $\textit{Cat}$, $f_{echo}$, and $f_{gru}$ denote the concatenation, echo encoder, and GRU operation, respectively. $A_{t}$ represents the candidate actions from the action space. We sample an action $a_{t}$ from $A_{t}$ according to the policy's predicted probability distribution.

Depth contains spatial information relating to the distance of the scene surfaces, which significantly facilitates embodied agent to avoid obstacles and navigate in 3D environment. We discussed that echoes can overcome and extend the limited RGB FoV to predict the depth of a wide FoV (Section~\ref{sec:depth_av}). The estimated depth of a wide FoV can be applied for better 3D embodied navigation when original depth is not available.

AudioGoal navigation~\cite{chen2020soundspaces,gan2020look} has recently been proposed to make the navigation task easier by introducing a sounding point goal, which serves as additional information of relative distance and angle to the agent. PointGoal navigation is a more challenging task. Unlike existing PointGoal visual navigation methods that heavily rely on visual observations~\cite{chen2020soundspaces,chaplot2020learning} (e.g. raw RGB or depth images), our approach experiments on utilizing the holistic spatial cues within binaural echoes for the PointGoal navigation. We also study to better navigate with audio-visual learning by leveraging echoes to visual observations. A neural network with three convolutional layers and a fully connected layer (each layer follows a \textit{ReLU} operation) is applied to process the visual observation (not visible in Fig.~\ref{fig:arch_nav}).

\setlength{\tabcolsep}{0.2pt}
\begin{table}[t]
    \centering
    \caption{Comparing depth estimation performance inside RGB FoV between our proposed models and baseline methods using Replica~\cite{straub2019replica} and Matterport3D~\cite{chang2017matterport3d} datasets.}
    \label{table:sota}
    \begin{tabular}{l|llllllll}
        \hline\noalign{\smallskip}
        Dataset & Method & RMSE($\downarrow$) & REL($\downarrow$) & Log10($\downarrow$) & $\delta_{1.25}$($\uparrow$) & $\delta_{1.25^{2}}$($\uparrow$) & $\delta_{1.25^{3}}$($\uparrow$)\\
        \noalign{\smallskip}
        \hline
        \multirow{7}{*}{Replica} &
        Average\cite{gao2020visualechoes} & 1.070 & 0.791 & 0.230 & 0.235& 0.509 & 0.750 \\
        & Echo2Depth\cite{gao2020visualechoes} & 0.969 & 0.753 & 0.204 & 0.441 & 0.631 & 0.752 \\
        & RGB2Depth\cite{gao2020visualechoes} & 0.374 & 0.202 & 0.076 & 0.749 & 0.883 & 0.945 \\
        & VisualEchoes\cite{gao2020visualechoes} & 0.346 & 0.172 & 0.068 & 0.798 & 0.905 & 0.950 \\
        & Materials\cite{parida2021beyond} & 0.249 & 0.118 & 0.046 & 0.869 & 0.943 & 0.970 \\
        & Ours(Echoes) & 0.797 & 0.534 & 0.171 & 0.544 & 0.708 & 0.802 \\
        & Ours(Echoes+RGB) & 0.294 & 0.166 & 0.060 & 0.814 & 0.912 & 0.958 \\
        \hline
        \multirow{7}{*}{Matterport3D} &
        Average\cite{gao2020visualechoes} & 1.913 & 0.714 & 0.237 & 0.264 & 0.538 & 0.697 \\
        & Echo2Depth\cite{gao2020visualechoes} & 1.778 & 0.507 & 0.192 & 0.464 & 0.642 & 0.759 \\
        & RGB2Depth\cite{gao2020visualechoes} & 1.090 & 0.260 & 0.111 & 0.592 & 0.802 & 0.910 \\
        & VisualEchoes\cite{gao2020visualechoes} & 0.998 & 0.193 & 0.083 & 0.711 & 0.878 & 0.945 \\
        & Materials\cite{parida2021beyond} & 0.950 & 0.175 & 0.079 & 0.733 & 0.886 & 0.948 \\
        & Ours(Echoes) & 1.535 & 0.465 & 0.184 & 0.476 & 0.664 & 0.781 \\
        & Ours(Echoes+RGB) & 0.777 & 0.161 & 0.069 & 0.775 & 0.874 & 0.943 \\
        \hline
    \end{tabular}
\end{table}
\setlength{\tabcolsep}{0.2pt}

\begin{figure}[!tbp]
    \centering
    \includegraphics[width=1.0\textwidth,keepaspectratio]{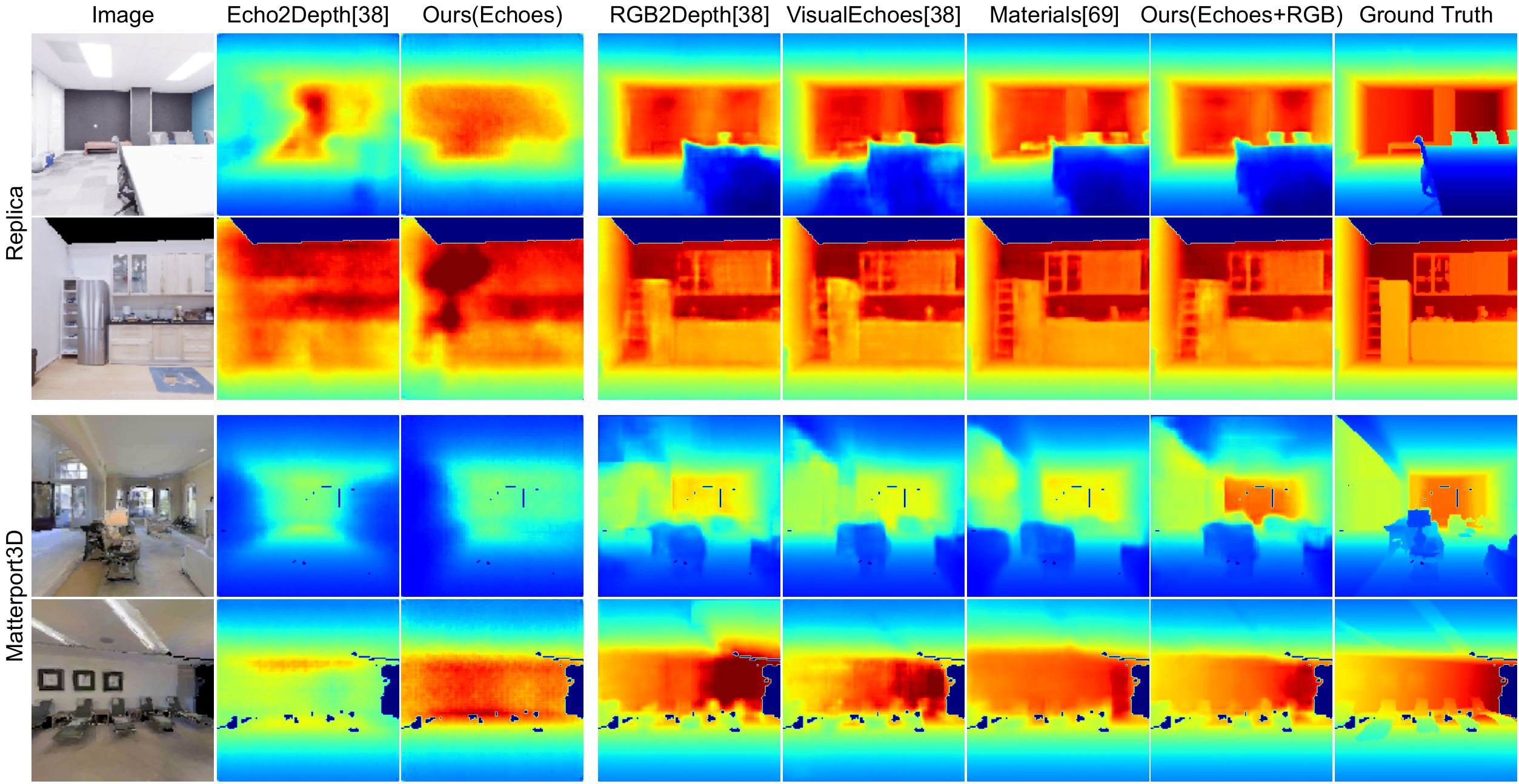}
    \caption{Visualization of depth prediction in comparison to baselines. The depth from second and third columns are estimated from using only echoes.}
    \label{fig:vis_depth}
\end{figure}

\section{Experiments}
\label{sec:exp}


In this section, we start by comparing our proposed model to previous state-of-the-art methods on depth prediction inside RGB FoV. Then, we report the experiments results of extending wide FoV depth prediction using echoes and RGB and applying them for embodied 3d navigation. For the task of depth prediction, we report the standard metrics of root mean squared error (RMSE), mean relative error (REL), mean log10 error (log10), and the thresholded accuracy of $\{\delta_{1.25}, \delta_{1.25^{2}}, \delta_{1.25^{3}}\}$~\cite{eigen2014depth,hu2019revisiting}. For navigation, we evaluate with the success rate normalized by inverse path length (SPL)~\cite{anderson2018evaluation}.

\subsection{Datasets and Echo Simulation}
\label{sec:data_sim}

SoundSpaces~\cite{chen2020soundspaces} is a realistic acoustic simulation platform, which augments the Habitat simulator~\cite{savva2019habitat}. Habitat~\cite{savva2019habitat} is an open-source 3D simulator that supports fast rendering for multiple datasets on RGB, depth, and semantics. SoundSpaces~\cite{chen2020soundspaces} enables audio rendering based on geometrical acoustic simulations for two sets of publicly available 3D environments Replica~\cite{straub2019replica} and Matterport3D~\cite{chang2017matterport3d}. Replica is a dataset with 3D meshes from real-world scans of 18 scenes of apartments, offices, hotels, and rooms. These scenes from Replica range in area from 9.5 to 141.5 $m^{2}$. The Matterport3D dataset consists of 85 large environments, range from 53.1 to 2921.3 $m^{2}$, which are real-world indoor environments, e.g. homes, with 3D meshes and image scans.

The SoundSpaces simulates acoustics by pre-computing Room Impulse Response (RIR) between a pair of sound source and microphone in the environment. The RIR is a transfer function between the source emitting and receiving pairs. It reflects the room geometry, materials, and the sound source location~\cite{kuttruff2016room}. Let $\mathcal{D} = \{(x^{i}, y^{i}, z^{i})\}_{i=1}^{N}$ denote the 1 to N navigable points in a scene of an environment, and let $\mathit{rir}^{i}$ denote the RIR at navigable point $\mathcal{D}^{i}$ (both the sound source and sound receiver are at the navigable point $\mathcal{D}^{i}$). The agent emits an omnidirectional sweep signal with a duration of 3ms~\cite{gao2020visualechoes} at each navigable location. At location $\mathcal{D}^{i}$, an ambisonic audio is generated by convolving the $\mathit{rir}^{i}$ with the sweep signal emitted from the same location. In order to perceive the spatial cues of a scene depending on the orientation from which to receive the signal, we convert the ambisonic audio into binaural echo responses $e^{i,j}$~\cite{zaunschirm2018binaural} by taking into account the shape of human ears and the head shadowing effect modeled in the binaural head- related transfer function (HRTF). $j \in \{0^{\circ}, 90^{\circ}, 180^{\circ}, 270^{\circ}\}$ denotes the orientation where the agent receives the binaural echoes.

\setlength{\tabcolsep}{1.0pt}
\begin{table}[t]
\centering
    \caption{Depth estimation using echoes received from multiple orientations through Replica~\cite{straub2019replica} and Matterport3D~\cite{chang2017matterport3d} datasets.}
    \label{table:echo}
    \begin{tabular}{c|cccccccc}
        \hline\noalign{\smallskip}
         Dataset & Echoes(pair) & RMSE($\downarrow$) & REL($\downarrow$) & Log10($\downarrow$) & $\delta_{1.25}$($\uparrow$) & $\delta_{1.25^{2}}$($\uparrow$) & $\delta_{1.25^{3}}$($\uparrow$)\\
        \noalign{\smallskip}
        \hline
        \multirow{2}{*}{Replica} &
        1 & 0.969 & 0.753 & 0.204 & 0.441 & 0.631 & 0.752 \\
        & 4 & 0.797 & 0.534 & 0.171 & 0.544 & 0.708 & 0.802 \\
        \noalign{\smallskip}
        \hline
        \multirow{2}{*}{Matterport3D} &
        1 & 1.778 & 0.507 & 0.192 & 0.464 & 0.642 & 0.759 \\
        & 4 & 1.535 & 0.465 & 0.184 & 0.476 & 0.664 & 0.781 \\
        \hline
    \end{tabular}
\end{table}
\setlength{\tabcolsep}{1.0pt}

\subsection{Depth Estimation from Echoes and RGB}
\label{sec:exp_depth}

\subsubsection{Comparison with state-of-the-art:}


We start by comparing our models against competitive baselines of Average, Echo2Depth, RGB2Depth, VisualEchoes~\cite{gao2020visualechoes}, and Materials~\cite{parida2021beyond} to estimate depth inside RGB FoV ($90^{\circ}$) in Table~\ref{table:sota}. We adopt same experimental setup as~\cite{gao2020visualechoes,parida2021beyond}. When using echoes alone, our method of combining echoes received from different orientations performs better than Echo2Depth~\cite{gao2020visualechoes}\footnote{We are not able to reproduce the result reported in paper~\cite{gao2020visualechoes}. Thus, we report our implemented result from following the description in the original paper.} with a large margin. With the presence of target orientation RGB image, our proposed approach achieves improvement of 15.0\% (Replica) and 22.1\% (Matterport3D) over VisualEchoes~\cite{gao2020visualechoes}. Furthermore, we observe that the method Materials~\cite{parida2021beyond} performs the best on Replica dataset while only attaining similar results as VisualEchoes~\cite{gao2020visualechoes} on Matterport3D. This may explain that the material cues brought from the pretrained material approach~\cite{bell2015material} has dominant impact for depth prediction on Replica. However, for large Matterport3D environment scene, its influence declines. Remarkably, our model achieves the state-of-the-art results on Matterport3D dataset, overwhelming Materials~\cite{parida2021beyond} around 18.2\% on RMSE. It is worth noting that the Materials~\cite{parida2021beyond} model has 316.9M parameters in comparison to our 21.7M. In addition, we show qualitative visualizations of depth estimation in Fig.~\ref{fig:vis_depth}. More qualitative examples are presented in supplementary materials. These indicate our proposed methods better perceive the geometrical information. 

Existing methods mainly focus on predicting depth inside RGB FoV where the RGB image is a strong cue and the improvement from leveraging echo is rather limited. Differently, in our work, we concentrate on leveraging echoes to perceive 3D structure outside the RGB FoV.

\begin{figure}[!tbp]
    \centering
    \includegraphics[width=1.0\textwidth,keepaspectratio]{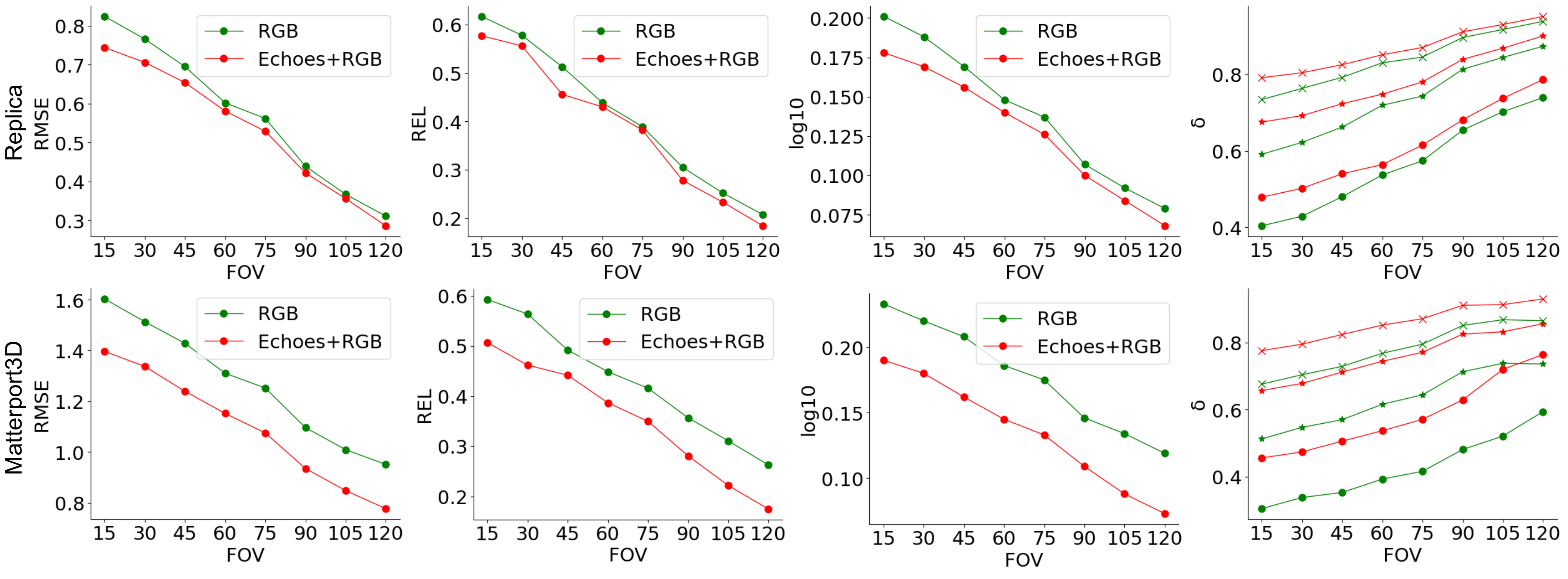}
    \caption{Depth prediction using RGB FoV $\in (0, 120]$, w/o echoes (green) and w/ echoes (red). For the thresholded accuracy $\delta$ (last column), the curves with $\bullet$, $\ast$, and x denote the $\delta_{1.25}$, $\delta_{1.25^{2}}$, and $\delta_{1.25^{3}}$, respectively.}
    \label{fig:curves}
\end{figure}

\subsubsection{Depth estimation from echoes: }

From Table~\ref{table:echo}, we find that combining echoes received from multiple orientations achieves better results than using one pair alone. Taking advantage of echoes from four different orientations achieves the best results of e.g. RMSE: 0.797 (Replica) and 1.535 (Matterport3D). The results tell that the echoes received from different orientation contain useful spatial information. We detail the performance difference among various echo orientation combinations in the supplementary materials. 

\subsubsection{Depth Estimation by Combining Echoes with RGB:}

We experiment our model in Fig.~\ref{fig:arch_depth} (w/o and w/ echoes) for depth extension (FoV $120^{\circ}$) using echoes and RGB of FoV $\in$ $\{15^{\circ}$, $30^{\circ}$, $45^{\circ}$, $60^{\circ}$, $75^{\circ}$, $90^{\circ}$, $105^{\circ}$, $120^{\circ}\}$. We observe from Fig.~\ref{fig:curves} that associating echoes outperforms the counterpart results (w/o echoes) over different FoV. Especially for Replica, the improvement gets smaller when increasing the RGB FoV. This indicates that the echoes serve as a strong spatial cue when goes to the region where RGB is not available. Thus, increasing the RGB FoV to apply to echoes does not bring large performance gain. Interestingly, when enlarging the RGB FoV for Matterport3D, the results of using echoes and RGB have relatively stable improvement than using RGB alone. This may because the Matterport3D contains large 3D environment scenes and the RGB image captures important geometric structure for large environment scenes.

\subsubsection{Leveraging echoes and RGB to predict depth of unseen areas:}

\begin{figure}[!tbp]
    \centering
    \includegraphics[width=1.0\textwidth,keepaspectratio]{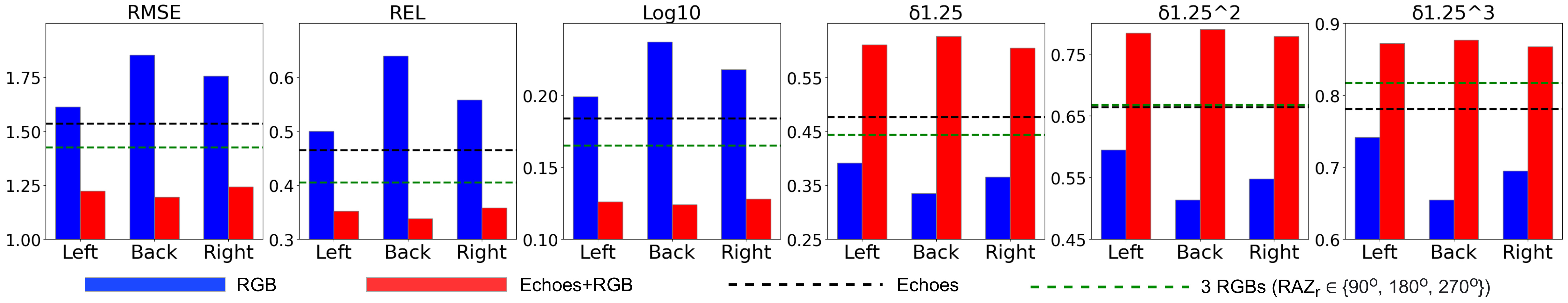}
    \caption{The bar charts of predicting ``Left'', ``Back'', and ``Right'' side depth from leveraging echoes to a forward RGB using Matterport3D.}
    \label{fig:bars}
\end{figure}

It is a more challenging problem when there is no overlap between the input RGB and target depth. Fig.~\ref{fig:bars} visualizes depth prediction metrics of the sideways (``left'' and ``right'') and ``back'' when giving echoes and RGB image. The input RGB and predicted depth are of FoV $90^{\circ}$. Using RGB image to infer depth of an unseen orientation may benefit from the similarity and extension of the visual surfaces. The visual similarity between the forward and backward is comparatively lower, thus it revels by the worse performance of the blue bar from ``Back'' depth prediction compared to the ``Left'' and ``Right''. The model of using echoes alone (dashed black line) performs better than using RGB image (blue bars). We also experiment depth estimation of a target orientation by using the RGB images from three rest orientations. For instance, we use the RGB images from ``Left'', ``Right'', and ``Back'' sides to predict the front depth. Its result is shown as dashed green line in Fig.~\ref{fig:bars}, which indicates investing additional cameras can bring performance gain but increase substantial computing complexity. 

However, fusing echoes into one RGB image (red bars) attains superior improvements. For all the metrics, the red bar surpasses the blue bar, blank dashed line, and the green dashed line by a large margin. These reflect the efficacy of fusing echoes into RGB image for exploiting the geometrical information. Specifically, we observe that, after fusing echoes into RGB, the performance differences predicting for among ``Left'', ``Right'', and ``Back'' get smaller for all the metrics. This is interesting because it suggests that the echoes capture complementary spatial information for each orientation, which also shows echoes contain strong geometrical cues when we go outside of the RGB FoV.

\subsection{Navigating Using Echoes and RGB}
\label{sec:exp_nav}

\subsubsection{Baselines:}
We consider three non-learning based and three visual sensing learning based baselines. The three non-learning baselines are Random, Rorward, and Goal follower~\cite{chen2019learning,savva2019habitat}. The three visual sensing based baselines predict an action from using the visual input of no visual observation, raw RGB image, and depth image. Similar to~\cite{chen2019learning,savva2019habitat,Chaplot2020LearningTE,gordon2019splitnet,kojima2019learn}, agents are allowed a time horizon of 500 actions for all tasks. More details are provided in the supplementary materials.

Table~\ref{tab:nav} summarizes the navigation performance of SPL in comparison with baseline methods using the test environments from Replica and Matterport3D datasets. Without using any sensory input, the non-learning baselines perform poorly for both datasets. Utilizing the GPS sensory input alone (Blind) boosts the performance SPL from 0.124 to 0.491 for Replica and from 0.197 to 0.425 for Matterport3D. Adding RGB sensor further improves the performance.


\setlength{\tabcolsep}{4pt}
\begin{table}[!tbp]
    \centering
        \caption{Navigation using estimated depth (italic rows) discussed in Section~\ref{sec:exp_depth} in comparison with baseline methods. The 90$^{\circ}$ or 120$^{\circ}$ represent the RGB or depth FoV.}
    \label{tab:nav}
    \begin{tabular}{lcc}
        \hline
        \noalign{\smallskip}
        Methods & Replica (SPL$\uparrow$) & Matterport3D (SPL$\uparrow$) \\
        \noalign{\smallskip}
        \hline
        Random & 0.044 & 0.021 \\
        Forward & 0.063 & 0.025 \\
        Goal follower & 0.124 & 0.197 \\
        \hline
        Blind & 0.491 & 0.425 \\
        RGB-90$^{\circ}$ & 0.526 & 0.448 \\
        \hline
        \textit{Depth-90$^{\circ}$(RGB-90$^{\circ}$)} & 0.527 & 0.431 \\
        \textit{Depth-90$^{\circ}$(Echoes+RGB-90$^{\circ}$)} & 0.570 & 0.473 \\
        \textit{Depth-120$^{\circ}$(Echoes+RGB-90$^{\circ}$)} & \bf0.591 & \bf0.488 \\
        \hline
    \end{tabular}
\end{table}
\setlength{\tabcolsep}{4pt}


\begin{figure}[hbt!]
    \centering
    \includegraphics[width=0.8\textwidth,keepaspectratio]{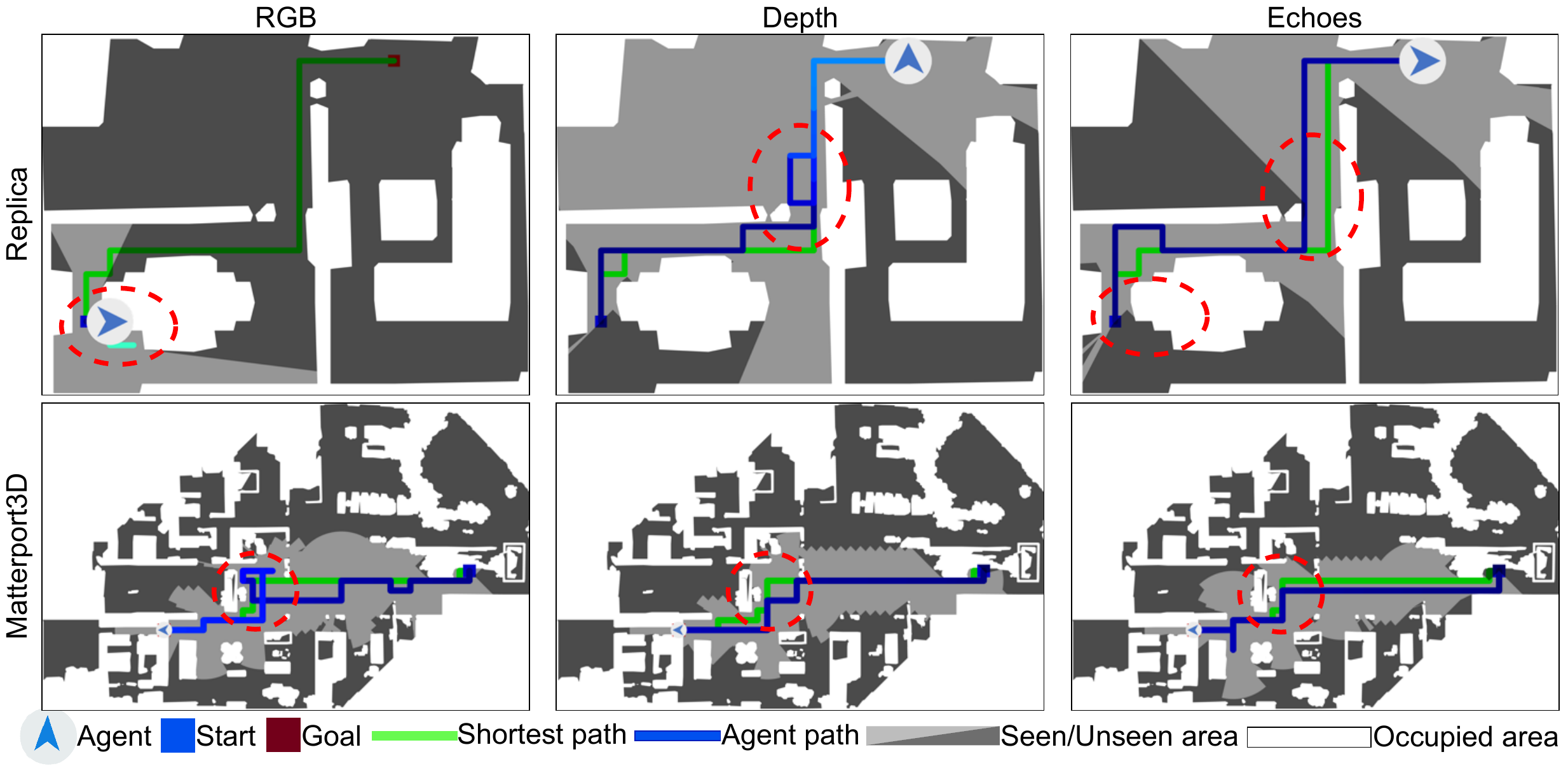}
    \caption{Navigation trajectories on top-down maps of using echoes in comparison with RGB and depth. Agent path fades from dark blue to light blue as time goes by.}
    \label{fig:nav}
\end{figure}

\subsubsection{Navigating using estimated wide field of view depth:} 
Depth is a strong cue for navigation. However, when the original depth is not available, we could apply the estimated depth from Section~\ref{sec:exp_depth} as input for the PointGoal navigation task. The \textit{Depth-90$^{\circ}$(RGB-90$^{\circ}$)} in Table~\ref{tab:nav} indicates a depth of FoV 90$^{\circ}$ estimated from RGB of FoV 90$^{\circ}$. We found the SPL of using \textit{Depth-90$^{\circ}$(RGB-90$^{\circ}$)} is fairly close to the result from using original RGB-90$^{\circ}$. Interestingly, \textit{Depth-90$^{\circ}$(Echoes+RGB-90$^{\circ}$)} obtains superior performance gain over the results from using original RGB-90$^{\circ}$ and \textit{Depth-90$^{\circ}$(RGB-90$^{\circ}$)}. Applying an estimated depth of wide FoV (e.g. 120$^{\circ}$) further improves the result. These suggest that the estimated depth, from using echoes and RGB, contains meaningful spatial cues and its wide FoV is beneficial for the PointGoal navigation. Thus, the depth estimated from echoes and RGB can be a good replacement when the original depth is not available. These findings motivate us to explore how the original echoes together with visual observations perform for the PointGoal navigation. 


\subsubsection{Navigating using binaural echoes:}
Compared to the results of using visual sensory input, the performance of directly utilizing echoes (Replica: 0.547 and Matterport3D: 0.474) in Table~\ref{tab:nav1} stands between using raw RGB image and using original depth. It revels that directly applying echoes captures stronger spatial cues than using raw RGB image for PointGoal navigation. These findings may result from i) the fact that echoes naturally capture holistic understanding of the 3D environment, thus have great spatial perception of the physical space; and ii) perceiving geometrical information of 3D environment to predict an action towards a point goal is technically a coarse-grained classification task for which echoes are a strong cue already. Fig.~\ref{fig:nav} shows examples of navigation trajectories on top-down maps using echoes in comparison with using raw RGB and original depth image. PointGoal RGB agent moves back and forth (light blue path) and bumps into obstacles multiple times. In contrast, the echoes, especially for the highlighted regions (dash red circles), better sense the obstacles and efficiently avoid backtracking. 


\subsubsection{Navigating using echoes and visual observation:}
In Table~\ref{tab:nav1}, combing echoes with the visual sensory input further improves the accuracy which indicating the efficacy of the audio-visual learning. The strong geometric structures contained in echoes and depth make the method of \textit{Echoes+Depth-90$^{\circ}$} perform better than \textit{Echoes} and \textit{Echoes+RGB-90$^{\circ}$}. In order to verify whether the models benefit from the holistic understanding of environment by echoes, we enlarge the FoV of input depth to fuse with echoes for navigation. When increasing the depth FoV from 90$^{\circ}$ to 120$^{\circ}$, the improvement gain from \textit{Echoes+Depth-120$^{\circ}$} over \textit{Depth-120$^{\circ}$} is smaller than the experiments of using FoV 90$^{\circ}$. This reflects our observation from depth estimation in Fig.~\ref{fig:curves}, \ref{fig:bars} and also indicates echoes can perceive strong geometrical cues when go outside the visual FoV.

\setlength{\tabcolsep}{4pt}
\begin{table}[!tbp]
    \centering
    \caption{Navigation performance from using echoes and leveraging echoes to vision.}
    \label{tab:nav1}
    \begin{tabular}{cccccc}
        \hline
        \noalign{\smallskip}
        Echoes & RGB-90$^{\circ}$ & Depth-90$^{\circ}$ & Depth-120$^{\circ}$ & Replica (SPL$\uparrow$) & Matterport3D (SPL$\uparrow$) \\
        \noalign{\smallskip}
        \hline
        \xmark & \cmark & \xmark & \xmark & 0.526 & 0.448 \\
        \cmark & \xmark & \xmark & \xmark & 0.547 & 0.474 \\
        \xmark & \xmark & \cmark & \xmark & 0.599 & 0.531 \\
        \hline
        \cmark & \cmark & \xmark & \xmark & 0.563 & 0.490 \\
        \cmark & \xmark & \cmark & \xmark & 0.613 & 0.553 \\
        \hline
        \xmark & \xmark & \xmark & \cmark & 0.624 & 0.546 \\
        \cmark & \xmark & \xmark & \cmark & 0.627 & 0.562 \\
        \hline
    \end{tabular}
\end{table}
\setlength{\tabcolsep}{4pt}

\section{Conclusion}

Our work sheds light on taking advantage of echoes to extend the geometrical understanding of physical space over visual observation. We leverage echoes received from multiple orientations to RGB for better depth estimation. Echoes can perceive a holistic understanding of the environment thus help to extend prediction outside the visual observation FoV. When the original depth is not available, a wide FoV depth that estimated from echeos and RGB better perceives the geometrical structure for embodied 3D navigation. Another important contribution of our work is we introduce PointGoal echo navigation, which outperforms PointGoal RGB navigation. Leveraging echoes to visual observation further improves the performance. These can facilitate future work in the field of embodied AI navigation.

\paragraph{\bf Acknowledgement} This work is supported by the Academy of Finland (projects 327910 \& 324346).

\clearpage
%
%
\bibliographystyle{splncs04}
\bibliography{main}
\end{document}